\documentclass[11pt,a4paper]{article}
\usepackage[hyperref]{acl2021}
\usepackage{times}
\usepackage{latexsym}

\usepackage{microtype}
\usepackage{graphicx}

\aclfinalcopy 

\title{Better than BERT but Worse than Baseline}
\author{Boxiang Liu \\
  Baidu Research \\
  \texttt{boxiangliu@baidu.com} \\\And
  Jiaji Huang \\
  Baidu Research \\
  \texttt{huangjiaji@baidu.com} \\\AND
  Xingyu Cai \\
  Baidu Research \\
  \texttt{xingyucai@baidu.com} \\\And
  Kenneth Church \\
  Baidu Research \\
  \texttt{kennethchurch@baidu.com}}

\date{}

\begin{document}
\maketitle
\begin{abstract}
This paper compares BERT-SQuAD and Ab3P on the \textbf{A}bbreviation \textbf{D}efinition \textbf{I}dentification (ADI) task.  ADI inputs a text and outputs short forms (abbreviations/acronyms) and long forms (expansions).
BERT with reranking improves over BERT without reranking but fails to reach the Ab3P rule-based baseline.
What is BERT missing?  Reranking introduces two new features: \textit{charmatch} and \textit{freq}.
The first feature identifies opportunities to take advantage of character constraints in acronyms and the second feature
identifies opportunities to take advantage of frequency constraints across documents.
\end{abstract}

\section{Introduction: Opportunities}

Transformers such as BERT \cite{devlin-etal-2019-bert} and ERNIE \cite{sun-etal-2019-baidu} have been extremely successful on a wide range of tasks.
Nevertheless, there are opportunities to improve BERT on numbers \cite{wallace2019nlp}, negation \cite{ettinger-2020-bert}, and more.

This paper compares BERT to Ab3P\footnote{\url{https://github.com/ncbi-nlp/Ab3P}} \cite{sohn2008abbreviation}
on the \textit{Abbreviation Definition Identification} (ADI) task described in Section~\ref{sec:ADI}.
ADI inputs texts and outputs pairs of short forms (SFs) and long forms (LFs).
A number of ADI systems were developed more than a decade ago; some use rules \cite{sohn2008abbreviation,schwartz2002simple} and others machine learning \cite{kuo2009bioadi}.

Section~\ref{sec:related_tasks} discusses similarities between ADI and question answering (QA). The QA dataset, SQuAD \cite{rajpurkar2016squad},
includes many types of questions, some of which are similar to ADI: \textit{What does X stand for}?  \textit{X} is a SF (abbreviation) and the answer is a LF (expansion).
A simple program based on BERT-SQuAD\footnote{https://huggingface.co/bert-large-uncased-whole-word-masking-finetuned-squad}
performs remarkably well on ADI benchmarks, though not as well as Ab3P, a strong rule-based baseline. 

Ab3P uses a set of 17 rules to extract SF-LF pairs. 
Rules were created iteratively.  Each iteration finds a rule that reduces the majority of missing cases.
The iteration stops when the desired recall has been achieved. 
One of these rules favors pairs with matching
characters.  That is, it is common for each character in an acronym to match each word in the expansion.

Why is BERT not doing better?  What is BERT missing?  Section~\ref{sec:features} uses a reranking approach
to improve BERT by adding two features that are easy to interpret: 
\begin{enumerate}
    \setlength{\itemsep}{0pt}
    \setlength{\parskip}{0pt}
    \setlength{\parsep}{0pt}
    \item \textit{charmatch} compares the first letter of LF to the first letter of SF, and
    \item \textit{freq} counts instances of ``LF (SF'' in a corpus of PubMed abstracts.
\end{enumerate}
It has been suggested that deep nets are
so powerful that feature engineering is no longer
necessary.  Reranking suggests this may not be correct, especially for tasks like ADI where
rule-based systems outperform BERT.

\subsection {The ADI Task}
\label{sec:ADI}
Abbreviations and acronyms are especially common in technical writing such as PubMed and arXiv\footnote{\url{https://arxiv.org/help/bulk_data}} \cite{veyseh-et-al-2020-what}, though they can be found in many other corpora such as Wikipedia.  The first time a SF is used in a paper, there is
usually a definition that connects the dots between the SF and the LF.  It is more common for definitions to parenthesize the SF, though both types of definitions are common:
\begin{enumerate}
    \setlength{\itemsep}{0pt}
    \setlength{\parskip}{0pt}
    \setlength{\parsep}{0pt}
    \item {\bf LF (SF)}: heat shock protein (HSP)
    \item {\bf SF (LF)}: HSP (heat shock protein)
\end{enumerate}

The ADI task takes a text as input, and outputs pairs of SFs and LFs that are defined in the input text. Four benchmarks have become standard in the ADI literature: Ab3P \cite{sohn2008abbreviation},
BIOADI \cite{kuo2009bioadi}, MEDSTRACT \cite{wren2005biomedical} and SH \cite{schwartz2002simple}.
Standard train, validation and test splits are available for download.\footnote{\url{http://bioc.sourceforge.net/BioCresources.html}}
All of these benchmarks are based on PubMed abstracts in ASCII without markup.\footnote{In other experiments,
we have run Ab3P on 2 TBs of arXiv papers, and found \LaTeX{} markup to be helpful because definitions often make use of italics.}

\subsection{Related Work, Tasks \& Tools}
\label{sec:related_tasks}

The ADI task is similar to a number of other tasks such as Question Answering (QA), Named Entity Recognition (NER), Acronym Disambiguation (AD), etc.
SQuAD \cite{rajpurkar2016squad} is a popular benchmark for QA systems.
SQuAD examples consist of questions, answers and contexts.  There are many types of questions, including some that are similar to ADI such as:
\begin{enumerate}
    \setlength{\itemsep}{0pt}
    \setlength{\parskip}{0pt}
    \setlength{\parsep}{0pt}
    \item Question: What does AFC stand for?
    \item Context: The \textit{American Football Conference} (AFC) champion Denver Broncos defeated the National Football Conference (NFC) champion Carolina Panthers 24–10 to earn their third Super Bowl title.
\end{enumerate}
Systems output a span, a substring of the context that answers the question such as: \textit{American Football Conference}.
This example suggests
BERT-SQuAD
may be a more modern
alternative to older rule-based approaches to ADI such as Ab3P.

Other studies have applied QA-like technology to a number of other tasks such as:
event extraction \cite{du-cardie-2020-event,liu2020event,feng2020probing,sun2020biomedical}, NER \cite{li-etal-2020-unified}, entity linking \cite{gu2021read}, coreference resolution \cite{wu2020corefqa}, and more. These studies suggest that a QA model with fine-tuning can achieve state-of-the-art (SOTA) results on a range of tasks. 

ADI is also similar to NER.  NER is used for a number of tasks that extract
spans, substrings of input text,
and labeling them with tags such as person, organization, location, etc. \cite{doddington2004automatic}, as
in ACE.\footnote{\url{https://www.ldc.upenn.edu/collaborations/past-projects/ace}}
Some benchmarks use BIO tags to label spans.  Each word in the input text is tagged as B (begins a span), I (inside a span) or O (otherwise).  Some benchmarks introduce tags such as B-disease and B-chemical to distinguish disease entities from chemical entities. \citet{lee2020biobert} show BioBERT is effective on a number of such NER benchmarks:\footnote{\url{https://github.com/dmis-lab/biobert-pytorch/tree/master/named-entity-recognition}} 
NCBI Disease \cite{dougan2014ncbi}
i2b2/VA \cite{uzuner20112010},
BC5CDR \cite{li2016biocreative},
BC4CHEMD \cite{krallinger2015chemdner},
BC2GM \cite{smith2008overview},
JNLPBA \cite{kim2004introduction},
LINNAEUS \cite{gerner2010linnaeus}
and Species-800 \cite{pafilis2013species}, as well as related tasks such as relation extraction and QA.



PubTator\footnote{\url{https://www.ncbi.nlm.nih.gov/research/pubtator/tutorial.html}} \cite{wei2012accelerating,wei2013pubtator,wei2019pubtator,leaman2016taggerone} is a practical NER tool.
PubTator annotations
for millions of PubMed abstracts are available for download.\footnote{\url{https://ftp.ncbi.nlm.nih.gov/pub/lu/PubTatorCentral/}}
PubTator identifies spans and tags them with six bioconcepts: genes, diseases, chemicals, mutations, species and cell lines.  PubTator also
links entities to ontologies such as MeSH.  PubTator links the gene p53, for example, to different points in the ontology for different species: humans, mice, fruit flies, etc. 


There have been concerns that work based on PubMed may not generalize well to other domains.
Benchmarks based on arXiv will be
used in competitions at AAAI-2021 for AI\footnote{\url{https://github.com/amirveyseh/AAAI-21-SDU-shared-task-1-AI}}
and AD\footnote{\url{https://github.com/amirveyseh/AAAI-21-SDU-shared-task-2-AD}} tasks \cite{veyseh-et-al-2020-what}.
The AI task uses NER-like labels with 5 tags:
B-short, B-long, I-short, I-long and O,
where B-short and I-short are used for SF spans, and B-long and I-long are used for LF spans.

This paper will use the ADI task which inputs definitions, as opposed
to AI and AD tasks which include subsequent mentions taken out of context.
Here is an example from the AI benchmark of a subsequent mention of MRC.\footnote{\textit{We also compare the performance of MRC cooperative NOMA with MRC
cooperative orthogonal multiple access (OMA), and we show that NOMA
has a better performance than OMA.}}
Using Google, we found the document.\footnote{\url{https://arxiv.org/abs/2003.07299}}
The definition of MRC
appears a few sentences earlier.

The AD task is based on word sense disambiguation (WSD).  The input sentence contains a SF, such as: \textit{The MRC technique employs a single copper loop of small radius both at the energy transmitter end and sensor node’s receiving end}.  The AD task is to choose the appropriate LF from a short list
of candidates:\footnote{Ambiguity is more common across documents than within documents.
MRC, for example, is unlikely to expand to two different LFs within the same document.
In this respect, SFs obey a so-called ``one sense per discourse'' constraint like word senses \cite{gale1992one}.
That is, the word \textit{bank} is ambiguous.  In one document, it may refer to a \textit{money bank}, and in another document, it may refer to a \textit{river bank},
but it is unlikely to be used both ways within the same document.}
\begin{enumerate}
    \setlength{\itemsep}{0pt}
    \setlength{\parskip}{0pt}
    \setlength{\parsep}{0pt}
    \item machine reading comprehension
    \item maximal ratio combining
    \item magnetic resonance coupling
\end{enumerate}
More context would be helpful.  With Google, we found the
document;\footnote{\url{https://arxiv.org/pdf/1805.07795.pdf}}
the definition, immediately before the input sentence,\footnote{\textit{The wireless transfer is performed by using the so called Magnetic Resonance Coupling (MRC) technique.}} resolves the ambiguity.


\begin{table}
\centering
\begin{tabular}{r | c c }
\textbf{Benchmark}   & \multicolumn{2}{|c}{\textbf{Method}} \\
            &	Ab3P            &  BERT-SQuAD     \\ \hline
Ab3P  	    & 	\textbf{0.889}  &  0.794	      \\
BIOADI	    & 	\textbf{0.838}  &  0.698	      \\
MEDSTRACT   & 	\textbf{0.943}  &  0.844	      \\
SH    	    & 	\textbf{0.858}  &  0.769	      \\ \hline
\end{tabular}
\caption{Ab3P has better F-scores on 4 benchmarks.}
\label{tab:BERT-SQuAD}
\end{table}

\begin{table}
\centering
\begin{tabular}{ r | r r }
\textbf{Benchmark}   & \multicolumn{2}{|c}{$Pr(correct|charmatch)$} \\
            &   not charmatch            &  charmatch     \\ \hline
Ab3P        &	0.15 & 0.96 \\
BIOADI      &	0.07 & 0.94 \\
MEDSTRACT   &   0.18 & 0.98 \\
SH          &   0.10 & 0.94      \\ \hline
\end{tabular}
\caption{BERT-SQuAD does not capture charmatch.}
\label{tab:BERT-SQuAD-charmatch}
\end{table}

\section{BERT-SQuAD: An Alternative to Ab3P for ADI}

The Ab3P method takes a text as input and outputs pairs of SFs and LFs that are defined in the input text.
BERT-SQuAD takes a question and context as input, and outputs a span from the document that answers the question.
For the comparisons in Table~\ref{tab:BERT-SQuAD}, we give BERT-SQuAD the SF from Ab3P output.
These SFs are turned into questions of the form: \textit{What does $<$SF$>$ stand for}?
Even with this unfair hint, BERT-SQuAD is less effective than Ab3P, as shown in Table~\ref{tab:BERT-SQuAD}.\footnote{BERT and Ab3P differ in many ways in addition to F-scores. BERT generalizes to many tasks, but it is bigger and slower and limited to inputs under 512 tokens.}

Many of the errors are ``off-by-one,'' where the candidate LF has one word too many or one too few, especially at the left edge of the LF.  The right edge tends to be easier because the
right edge of the LF is often delimited by a 
parenthesis between the LF and the SF.

\subsection{Two More Features: Charmatch \& Freq}
\label{sec:features}

What is BERT-SQuAD missing?  Consider the example: \textit{healthy controls (HC)}.  In this case, BERT-SQuAD drops the first word from the LF, returning \textit{controls} instead of \textit{healthy controls}.  BERT's candidate
violates a constraint on characters, where the bold characters in \underline{\textbf{\textit{h}}}ealthy \underline{\textbf{\textit{c}}}ontrols are likely to match the characters in the SF (HC).
Ab3P uses old-fashioned rules to capture this constraint.

\begin{table}
\centering
\begin{tabular}{ r | r  r  r r  }
\textbf{Rank}   & \multicolumn{4}{c}{\textbf{Benchmark}} \\
  & Ab3P  &	BIOADI & MESTRACT & SH \\ \hline
0  & 920 & 1129	& 128 &	 698 \\
1  &  85 &  103	&  11 &	  58 \\
2  &  19 &  124	&   3 &	  25 \\
3  &   8 &   68	&     &	   4 \\
4  &   6 &   33	&     &	     \\	\hline
\end{tabular}
\caption{\# correct by rank (position in n-best list)}
\label{tab:BERT-SQuAD-top5}
\end{table}

It appears this character constraint is missing from
BERT-SQuAD.
To test this hypothesis, 
we introduce a simple boolean feature, \textit{charmatch}, that compares the first character of the SF to the first character of
the candidate LF.  Table~\ref{tab:BERT-SQuAD-charmatch} shows that candidates from BERT-SQuAD are more likely to be correct when these characters match.

In addition to \textit{charmatch}, we identified another promising feature that we call \textit{freq}.
Consider the example: \textit{Latent herpes simplex virus (HSV) has been demonstrated in...}
Again, BERT-SQuAD is off by one, 
but this time, the candidate LF is too long: \textit{Latent herpes simplex virus}.
The freq feature takes advantage of the fact that many of these SFs are defined in thousands of PubMed abstracts.
The freq feature uses suffix arrays \cite{manber1993suffix} to count the number of matches of: SF + ` (' + LF in PubMed.
In this example, we found 6075 instances of ``herpes simplex virus (HSV'', but only
8 instances of ``Latent herpes simplex virus (HSV.''  Of course, raw frequencies need to be normalized
appropriately because shorter strings tend to be more frequent than longer strings.

\subsection{12 Models: Rank + Charmatch + Freq}

To side-step difficult normalization and feature combination questions, we make use of reranking and machine learning.
BERT-SQuAD was modified to output top-k candidates instead of just the top candidate.  Table~\ref{tab:BERT-SQuAD-top5} shows 
there are more correct candidates in top position (rank 0), but there are also many correct candidates in other
positions.

\begin{table}
\centering
\begin{tabular}{ r | r r r r r r  }
\textbf{Coef}   & \multicolumn{6}{| c}{\textbf{Model}} \\
  & 1 &	2 & 3 &	4 & 5 &	6  \\	\hline
$\beta_0$ & 1.6  &  0.7 &  1.9 & 1.4  & -1.2 & -2.5 \\
$\beta_1$ & -3.3 & -1.6 & -3.9 & -3.3 & -3.2 & -1.5 \\
$\beta_2$ & 0 & 0 & 0 & 0 & 3.5 & 3.8  \\
$\beta_3$ & 0 & 0 & 0 & 0 & 0 & 0 \\
  & \multicolumn{6}{| c}{\textbf{Model}} \\
  &  7 &	8 & 9 &	10 & 11	& 12 \\	\hline
$\beta_0$ & -1.0 & -1.9 & -2.7 & -5.2 & -3.2 & -3.1 \\
$\beta_1$ & -4.0 & -3.2 & -2.9 & -1.5 & -3.8 & -2.9 \\
$\beta_2$ &  3.9 & 4.1 & 3.7 & 5.2 & 4.7 & 4.3 \\
$\beta_3$ & 0 & 0 & 0.3 & 0.5 & 0.4 & 0.3 \\ \hline
\end{tabular}
\caption{Coefficients for 12 logistic regression models: $z = \beta_0 + \beta_1 ~rank + \beta_2~charmatch + \beta_3~log(1+freq)$}
\label{tab:coef}
\end{table}

\begin{table}
\centering
\begin{tabular}{ c | c c c c }
\textbf{Features} &	\multicolumn{4}{| c}{\textbf{\textbf{Benchmark}}} \\
 & Ab3P & BIOADI & MEDS. & SH \\ \hline
1 (M 1-4) & 0.829 & 0.660 & 0.865 & 0.796 \\
2 (M 5-8) & 0.908 & 0.787 & 0.945 & 0.901 \\
3 (M 9-12) & 0.936 & 0.948 & 0.975 & 0.949  \\ \hline
\end{tabular}
\caption{Models with more features are more confident when they are correct.
Models 1-4 use a single feature (rank).  Models 5-8 add charmatch.  Models 9-12 add
freq.  Confidence is estimated as median $\sigma(z)$ for correct candidates.}
\label{tab:confidence}
\end{table}

A dozen logistic regression models are used to rerank the top 5 candidates.   Models  1-4 use eq~(\ref{eqn:rerank1}), models 5-8 use eq~(\ref{eqn:rerank2}) and models 9-12 use eq~(\ref{eqn:rerank3}).  $y$ comes from the 4 gold sets: models
1,5 \& 9 use the Ab3P benchmark for $y$, models 2, 6 \& 10 use BIOADI,  models 3, 7 \& 11 use MEDSTRACT, and models
4, 8 \& 12 use SH.  Coefficients are shown in Table~\ref{tab:coef}.  All coefficients are significant.  Reranking sorts candidates by $z$, as defined in Table~\ref{tab:coef}.
\setlength{\belowdisplayskip}{1pt} \setlength{\belowdisplayshortskip}{1pt}
\setlength{\abovedisplayskip}{1pt} \setlength{\abovedisplayshortskip}{1pt}
\begin{equation}
    y \sim rank
    \label{eqn:rerank1}
\end{equation}
\begin{equation}
    y \sim rank + charmatch
    \label{eqn:rerank2}
\end{equation}
\begin{equation}
    y \sim rank + charmatch + log(1+freq)
    \label{eqn:rerank3}
\end{equation}


\begin{figure}[ht!]
\centering
\includegraphics[width=0.5\textwidth]{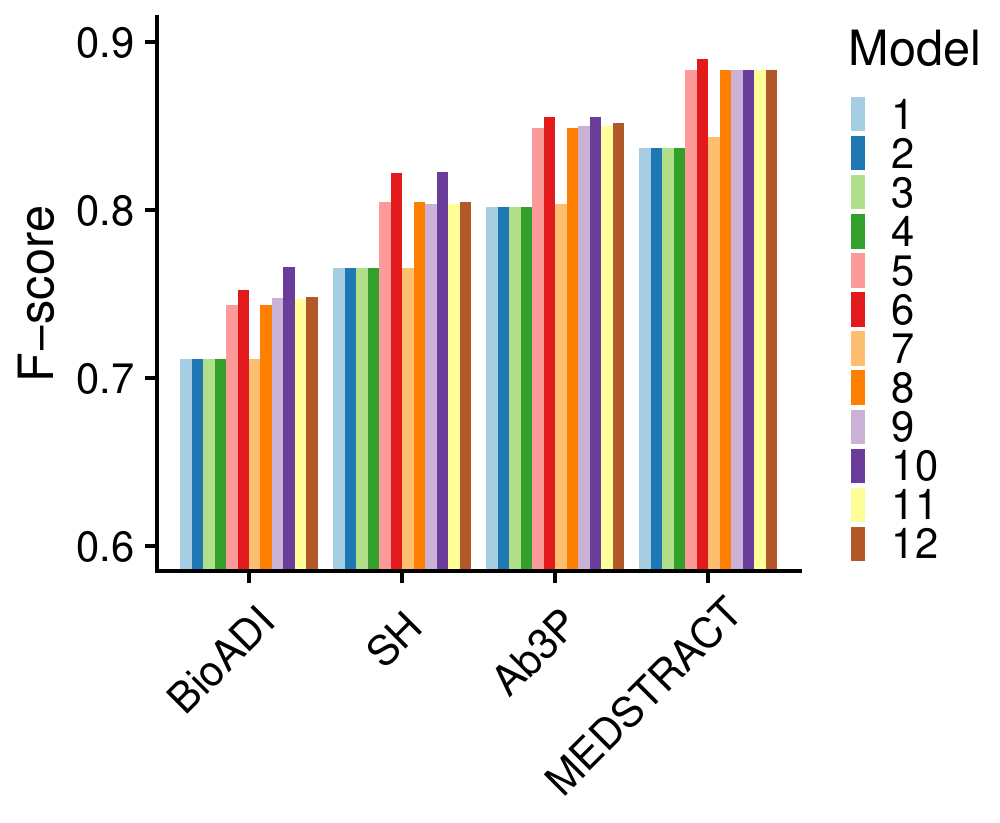}
\caption{F-score by model and benchmark}
\label{fig:rerank_pivot}
\end{figure}

\begin{figure}[ht!]
\centering
\includegraphics[width=0.4\textwidth]{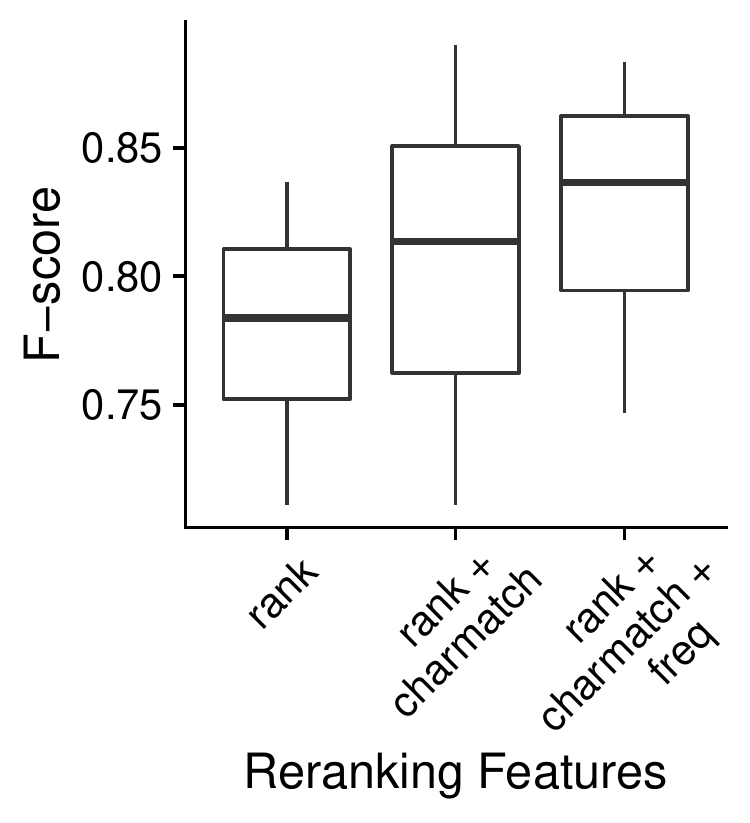}
\caption{More features increase F-scores.}
\label{fig:rerank}
\end{figure}

\section{Results}

Scores for 12 BERT-SQuAD models and 4 benchmarks are shown in Figure~\ref{fig:rerank_pivot}.  These results
are summarized in Figure~\ref{fig:rerank}.  Adding more features to BERT-SQuAD improves F-scores.  Table~\ref{tab:confidence} shows that models with more features are more confident.

\section{Conclusions}

This paper compared BERT-SQuAD to a rule-based baseline, Ab3P, on the ADI task.
We proposed a reranking improvement to BERT that takes advantage of two features: charmatch
and freq.  F-scores for the proposed solution are better than BERT but worse than baseline,
suggesting the two features shed light on opportunities for improving BERT-like models.

\bibliographystyle{acl_natbib}
\bibliography{anthology,acl2021}

\begin{thebibliography}{33}
\expandafter\ifx\csname natexlab\endcsname\relax\def\natexlab#1{#1}\fi

\bibitem[{Devlin et~al.(2019)Devlin, Chang, Lee, and
  Toutanova}]{devlin-etal-2019-bert}
Jacob Devlin, Ming-Wei Chang, Kenton Lee, and Kristina Toutanova. 2019.
\newblock \href {https://doi.org/10.18653/v1/N19-1423} {{BERT}: Pre-training of
  deep bidirectional transformers for language understanding}.
\newblock In \emph{Proceedings of the 2019 Conference of the North {A}merican
  Chapter of the Association for Computational Linguistics: Human Language
  Technologies, Volume 1 (Long and Short Papers)}, pages 4171--4186,
  Minneapolis, Minnesota. Association for Computational Linguistics.

\bibitem[{Doddington et~al.(2004)Doddington, Mitchell, Przybocki, Ramshaw,
  Strassel, and Weischedel}]{doddington2004automatic}
George~R Doddington, Alexis Mitchell, Mark~A Przybocki, Lance~A Ramshaw,
  Stephanie~M Strassel, and Ralph~M Weischedel. 2004.
\newblock The automatic content extraction (ace) program-tasks, data, and
  evaluation.
\newblock In \emph{Lrec}, volume~2, pages 837--840. Lisbon.

\bibitem[{Do{\u{g}}an et~al.(2014)Do{\u{g}}an, Leaman, and Lu}]{dougan2014ncbi}
Rezarta~Islamaj Do{\u{g}}an, Robert Leaman, and Zhiyong Lu. 2014.
\newblock Ncbi disease corpus: a resource for disease name recognition and
  concept normalization.
\newblock \emph{Journal of biomedical informatics}, 47:1--10.

\bibitem[{Du and Cardie(2020)}]{du-cardie-2020-event}
Xinya Du and Claire Cardie. 2020.
\newblock \href {https://doi.org/10.18653/v1/2020.emnlp-main.49} {Event
  extraction by answering (almost) natural questions}.
\newblock In \emph{Proceedings of the 2020 Conference on Empirical Methods in
  Natural Language Processing (EMNLP)}, pages 671--683, Online. Association for
  Computational Linguistics.

\bibitem[{Ettinger(2020)}]{ettinger-2020-bert}
Allyson Ettinger. 2020.
\newblock \href {https://doi.org/10.1162/tacl_a_00298} {What {BERT} is not:
  Lessons from a new suite of psycholinguistic diagnostics for language
  models}.
\newblock \emph{Transactions of the Association for Computational Linguistics},
  8:34--48.

\bibitem[{Feng et~al.(2020)Feng, Yuan, and Zhang}]{feng2020probing}
Rui Feng, Jie Yuan, and Chao Zhang. 2020.
\newblock Probing and fine-tuning reading comprehension models for few-shot
  event extraction.
\newblock \emph{arXiv preprint arXiv:2010.11325}.

\bibitem[{Gale et~al.(1992)Gale, Church, and Yarowsky}]{gale1992one}
William~A Gale, Kenneth Church, and David Yarowsky. 1992.
\newblock One sense per discourse.
\newblock In \emph{Speech and Natural Language: Proceedings of a Workshop Held
  at Harriman, New York, February 23-26, 1992}.

\bibitem[{Gerner et~al.(2010)Gerner, Nenadic, and Bergman}]{gerner2010linnaeus}
Martin Gerner, Goran Nenadic, and Casey~M Bergman. 2010.
\newblock Linnaeus: a species name identification system for biomedical
  literature.
\newblock \emph{BMC bioinformatics}, 11(1):1--17.

\bibitem[{Gu et~al.(2021)Gu, Qu, Wang, Huai, Yuan, and Gui}]{gu2021read}
Yingjie Gu, Xiaoye Qu, Zhefeng Wang, Baoxing Huai, Nicholas~Jing Yuan, and
  Xiaolin Gui. 2021.
\newblock Read, retrospect, select: An mrc framework to short text entity
  linking.
\newblock \emph{arXiv preprint arXiv:2101.02394}.

\bibitem[{Kim et~al.(2004)Kim, Ohta, Tsuruoka, Tateisi, and
  Collier}]{kim2004introduction}
Jin-Dong Kim, Tomoko Ohta, Yoshimasa Tsuruoka, Yuka Tateisi, and Nigel Collier.
  2004.
\newblock Introduction to the bio-entity recognition task at jnlpba.
\newblock In \emph{Proceedings of the international joint workshop on natural
  language processing in biomedicine and its applications}, pages 70--75.
  Citeseer.

\bibitem[{Krallinger et~al.(2015)Krallinger, Rabal, Leitner, Vazquez, Salgado,
  Lu, Leaman, Lu, Ji, Lowe et~al.}]{krallinger2015chemdner}
Martin Krallinger, Obdulia Rabal, Florian Leitner, Miguel Vazquez, David
  Salgado, Zhiyong Lu, Robert Leaman, Yanan Lu, Donghong Ji, Daniel~M Lowe,
  et~al. 2015.
\newblock The chemdner corpus of chemicals and drugs and its annotation
  principles.
\newblock \emph{Journal of cheminformatics}, 7(1):1--17.

\bibitem[{Kuo et~al.(2009)Kuo, Ling, Lin, and Hsu}]{kuo2009bioadi}
Cheng-Ju Kuo, Maurice~HT Ling, Kuan-Ting Lin, and Chun-Nan Hsu. 2009.
\newblock Bioadi: a machine learning approach to identifying abbreviations and
  definitions in biological literature.
\newblock In \emph{BMC bioinformatics}, volume~10, pages 1--10. BioMed Central.

\bibitem[{Leaman and Lu(2016)}]{leaman2016taggerone}
Robert Leaman and Zhiyong Lu. 2016.
\newblock Taggerone: joint named entity recognition and normalization with
  semi-markov models.
\newblock \emph{Bioinformatics}, 32(18):2839--2846.

\bibitem[{Lee et~al.(2020)Lee, Yoon, Kim, Kim, Kim, So, and
  Kang}]{lee2020biobert}
Jinhyuk Lee, Wonjin Yoon, Sungdong Kim, Donghyeon Kim, Sunkyu Kim, Chan~Ho So,
  and Jaewoo Kang. 2020.
\newblock Biobert: a pre-trained biomedical language representation model for
  biomedical text mining.
\newblock \emph{Bioinformatics}, 36(4):1234--1240.

\bibitem[{Li et~al.(2016)Li, Sun, Johnson, Sciaky, Wei, Leaman, Davis,
  Mattingly, Wiegers, and Lu}]{li2016biocreative}
Jiao Li, Yueping Sun, Robin~J Johnson, Daniela Sciaky, Chih-Hsuan Wei, Robert
  Leaman, Allan~Peter Davis, Carolyn~J Mattingly, Thomas~C Wiegers, and Zhiyong
  Lu. 2016.
\newblock Biocreative v cdr task corpus: a resource for chemical disease
  relation extraction.
\newblock \emph{Database}, 2016.

\bibitem[{Li et~al.(2020)Li, Feng, Meng, Han, Wu, and
  Li}]{li-etal-2020-unified}
Xiaoya Li, Jingrong Feng, Yuxian Meng, Qinghong Han, Fei Wu, and Jiwei Li.
  2020.
\newblock \href {https://doi.org/10.18653/v1/2020.acl-main.519} {A unified
  {MRC} framework for named entity recognition}.
\newblock In \emph{Proceedings of the 58th Annual Meeting of the Association
  for Computational Linguistics}, pages 5849--5859, Online. Association for
  Computational Linguistics.

\bibitem[{Liu et~al.(2020)Liu, Chen, Liu, Bi, and Liu}]{liu2020event}
Jian Liu, Yubo Chen, Kang Liu, Wei Bi, and Xiaojiang Liu. 2020.
\newblock Event extraction as machine reading comprehension.
\newblock In \emph{Proceedings of the 2020 Conference on Empirical Methods in
  Natural Language Processing (EMNLP)}, pages 1641--1651.

\bibitem[{Manber and Myers(1993)}]{manber1993suffix}
Udi Manber and Gene Myers. 1993.
\newblock Suffix arrays: a new method for on-line string searches.
\newblock \emph{siam Journal on Computing}, 22(5):935--948.

\bibitem[{Pafilis et~al.(2013)Pafilis, Frankild, Fanini, Faulwetter, Pavloudi,
  Vasileiadou, Arvanitidis, and Jensen}]{pafilis2013species}
Evangelos Pafilis, Sune~P Frankild, Lucia Fanini, Sarah Faulwetter, Christina
  Pavloudi, Aikaterini Vasileiadou, Christos Arvanitidis, and Lars~Juhl Jensen.
  2013.
\newblock The species and organisms resources for fast and accurate
  identification of taxonomic names in text.
\newblock \emph{PloS one}, 8(6):e65390.

\bibitem[{Rajpurkar et~al.(2016)Rajpurkar, Zhang, Lopyrev, and
  Liang}]{rajpurkar2016squad}
Pranav Rajpurkar, Jian Zhang, Konstantin Lopyrev, and Percy Liang. 2016.
\newblock Squad: 100,000+ questions for machine comprehension of text.
\newblock In \emph{Proceedings of the 2016 Conference on Empirical Methods in
  Natural Language Processing}, pages 2383--2392.

\bibitem[{Schwartz and Hearst(2002)}]{schwartz2002simple}
Ariel~S Schwartz and Marti~A Hearst. 2002.
\newblock A simple algorithm for identifying abbreviation definitions in
  biomedical text.
\newblock In \emph{Biocomputing 2003}, pages 451--462. World Scientific.

\bibitem[{Smith et~al.(2008)Smith, Tanabe, nee Ando, Kuo, Chung, Hsu, Lin,
  Klinger, Friedrich, Ganchev et~al.}]{smith2008overview}
Larry Smith, Lorraine~K Tanabe, Rie~Johnson nee Ando, Cheng-Ju Kuo, I-Fang
  Chung, Chun-Nan Hsu, Yu-Shi Lin, Roman Klinger, Christoph~M Friedrich, Kuzman
  Ganchev, et~al. 2008.
\newblock Overview of biocreative ii gene mention recognition.
\newblock \emph{Genome biology}, 9(2):1--19.

\bibitem[{Sohn et~al.(2008)Sohn, Comeau, Kim, and
  Wilbur}]{sohn2008abbreviation}
Sunghwan Sohn, Donald~C Comeau, Won Kim, and W~John Wilbur. 2008.
\newblock Abbreviation definition identification based on automatic precision
  estimates.
\newblock \emph{BMC bioinformatics}, 9(1):402.

\bibitem[{Sun et~al.(2020)Sun, Yang, Wang, Zhang, Lin, and
  Wang}]{sun2020biomedical}
Cong Sun, Zhihao Yang, Lei Wang, Yin Zhang, Hongfei Lin, and Jian Wang. 2020.
\newblock Biomedical named entity recognition using bert in the machine reading
  comprehension framework.
\newblock \emph{arXiv preprint arXiv:2009.01560}.

\bibitem[{Sun et~al.(2019)Sun, Jiang, Xiong, He, Wu, and
  Wang}]{sun-etal-2019-baidu}
Meng Sun, Bojian Jiang, Hao Xiong, Zhongjun He, Hua Wu, and Haifeng Wang. 2019.
\newblock \href {https://doi.org/10.18653/v1/W19-5341} {{B}aidu neural machine
  translation systems for {WMT}19}.
\newblock In \emph{Proceedings of the Fourth Conference on Machine Translation
  (Volume 2: Shared Task Papers, Day 1)}, pages 374--381, Florence, Italy.
  Association for Computational Linguistics.

\bibitem[{Uzuner et~al.(2011)Uzuner, South, Shen, and DuVall}]{uzuner20112010}
{\"O}zlem Uzuner, Brett~R South, Shuying Shen, and Scott~L DuVall. 2011.
\newblock 2010 i2b2/va challenge on concepts, assertions, and relations in
  clinical text.
\newblock \emph{Journal of the American Medical Informatics Association},
  18(5):552--556.

\bibitem[{Veyseh et~al.(2020)Veyseh, Dernoncourt, Tran, and
  Nguyen}]{veyseh-et-al-2020-what}
Amir Pouran~Ben Veyseh, Franck Dernoncourt, Quan~Hung Tran, and Thien~Huu
  Nguyen. 2020.
\newblock {What Does This Acronym Mean? Introducing a New Dataset for Acronym
  Identification and Disambiguation}.
\newblock In \emph{Proceedings of COLING}.

\bibitem[{Wallace et~al.(2019)Wallace, Wang, Li, Singh, and
  Gardner}]{wallace2019nlp}
Eric Wallace, Yizhong Wang, Sujian Li, Sameer Singh, and Matt Gardner. 2019.
\newblock Do nlp models know numbers? probing numeracy in embeddings.
\newblock In \emph{Proceedings of the 2019 Conference on Empirical Methods in
  Natural Language Processing and the 9th International Joint Conference on
  Natural Language Processing (EMNLP-IJCNLP)}, pages 5310--5318.

\bibitem[{Wei et~al.(2019)Wei, Allot, Leaman, and Lu}]{wei2019pubtator}
Chih-Hsuan Wei, Alexis Allot, Robert Leaman, and Zhiyong Lu. 2019.
\newblock Pubtator central: automated concept annotation for biomedical full
  text articles.
\newblock \emph{Nucleic acids research}, 47(W1):W587--W593.

\bibitem[{Wei et~al.(2012)Wei, Harris, Li, Berardini, Huala, Kao, and
  Lu}]{wei2012accelerating}
Chih-Hsuan Wei, Bethany~R Harris, Donghui Li, Tanya~Z Berardini, Eva Huala,
  Hung-Yu Kao, and Zhiyong Lu. 2012.
\newblock Accelerating literature curation with text-mining tools: a case study
  of using pubtator to curate genes in pubmed abstracts.
\newblock \emph{Database}, 2012.

\bibitem[{Wei et~al.(2013)Wei, Kao, and Lu}]{wei2013pubtator}
Chih-Hsuan Wei, Hung-Yu Kao, and Zhiyong Lu. 2013.
\newblock Pubtator: a web-based text mining tool for assisting biocuration.
\newblock \emph{Nucleic acids research}, 41(W1):W518--W522.

\bibitem[{Wren et~al.(2005)Wren, Chang, Pustejovsky, Adar, Garner, and
  Altman}]{wren2005biomedical}
Jonathan~D Wren, Jeffrey~T Chang, James Pustejovsky, Eytan Adar, Harold~R
  Garner, and Russ~B Altman. 2005.
\newblock Biomedical term mapping databases.
\newblock \emph{Nucleic acids research}, 33(suppl\_1):D289--D293.

\bibitem[{Wu et~al.(2020)Wu, Wang, Yuan, Wu, and Li}]{wu2020corefqa}
Wei Wu, Fei Wang, Arianna Yuan, Fei Wu, and Jiwei Li. 2020.
\newblock Corefqa: Coreference resolution as query-based span prediction.
\newblock In \emph{Proceedings of the 58th Annual Meeting of the Association
  for Computational Linguistics}, pages 6953--6963.

\end{thebibliography}


\end{document}